\documentclass[10pt,twocolumn,letterpaper]{article}

\usepackage{cvpr}
\usepackage{times}
\usepackage{epsfig}
\usepackage{graphicx}
\usepackage{amsmath}
\usepackage{amssymb}

\usepackage{verbatim}
\usepackage{float}

\graphicspath{{pic/}}

\usepackage[pagebackref=true,breaklinks=true,letterpaper=true,colorlinks,bookmarks=false]{hyperref}

\cvprfinalcopy % *** Uncomment this line for the final submission

\def\cvprPaperID{0} % *** Enter the CVPR Paper ID here

\ifcvprfinal\fi
\begin{document}

%%%%%%%%% TITLE
%\title{Convolutional Context Network for Video Temporal Modeling}
%\title{Localizing Video Clips via Joint Language-Vision Representation}
%\title{Localizing Video Clips via Context Enhanced Convolutional Network}
%\title{Localizing Clips via Exploiting Language Augmented Video Context}
%\title{Localizing Clip of Interest via Language Augmented Video Context Encoder}
%\title{Localizing Clip of Interest via Natural Language Description }
\title{Attentive Sequence to Sequence Translation for Localizing Clips of Interest\\ by Natural Language Descriptions}

\author{Ke Ning\thanks{This work was done while Ke Ning was visiting University of Technology Sydney}\\
Zhejiang University\\
{\tt\small ningke@zju.edu.cn}
\and
Linchao Zhu\\
University of Technology Sydney\\
{\tt\small zhulinchao7@gmail.com}
\and
Ming Cai\\
Zhejiang University\\
{\tt\small cm@zju.edu.cn}
\and
Yi Yang\\
University of Technology Sydney\\
{\tt\small Yi.Yang@uts.edu.au}
\and
Di Xie\\
Hikvision Research Institute\\
{\tt\small xiedi@hikvision.com}
\and
Fei Wu\\
Zhejiang University\\
{\tt\small wufei@zju.edu.cn}
}

\maketitle
\thispagestyle{empty}

\begin{comment}
\begin{figure*}[h!]
\begin{center}
\includegraphics[width=.95\linewidth]{Intro3.pdf}
%\def\svgwidth{0.95\linewidth}
%\input{pic/Introduction.pdf_tex}
\end{center}
\caption{An illustration of localizing clips of interest by natural language description.
    In this example, we are looking for the clip that the baby takes something out of his mouth. %MORE NEEDED?
The video clip above the yellow bar is the clip we are interested in.}
\label{fig:intro}
\vspace{-0.2cm}
\end{figure*}
\end{comment}

%%%%%%%%% ABSTRACT
\begin{abstract}
We propose a novel attentive sequence to sequence translator (ASST) for 
clip localization in videos by natural language descriptions. We make two contributions. 
First, we propose a bi-directional Recurrent Neural Network (RNN) with a finely calibrated vision-language attentive mechanism
to comprehensively understand the free-formed natural language descriptions. 
The RNN parses natural language descriptions in two directions, and the attentive model attends every meaningful word or phrase to each frame, thereby resulting in a more detailed understanding of video content and description semantics. 
Second, we design a hierarchical architecture for the network to jointly model language descriptions and video content. 
Given a video-description pair, the network generates a matrix representation, \ie, a sequence of vectors.
Each vector in the matrix represents a video frame conditioned by the description. 
The 2D representation not only preserves the temporal dependencies of frames but also provides an
effective way to perform frame-level video-language matching.
The hierarchical architecture 
exploits video content with multiple granularities, ranging from subtle details to
global context.
Integration of the multiple granularities yields a robust representation for multi-level video-language abstraction. 
We validate the effectiveness of our ASST on two large-scale datasets.
Our ASST outperforms the state-of-the-art by $4.28\%$ in Rank$@1$ on the DiDeMo dataset. On the Charades-STA dataset, we significantly improve the state-of-the-art by $13.41\%$ in Rank$@1,IoU=0.5$.
%Our method outperforms the state-of-the-art by $5.35\%$ in Rank$@1$ on the DiDeMo dataset. 
%Validating the standalone 2D representation on the action detection task,  our method obtains $15.3\%$ improvements on $mAP@IoU~0.5$ over the state-of-the-arts on THUMOS 14. The relative improvement is $51.3\%$.
\end{abstract}

%%%%%%%%% BODY TEXT

\begin{figure*}[h!]
\begin{center}
\includegraphics[width=.95\linewidth]{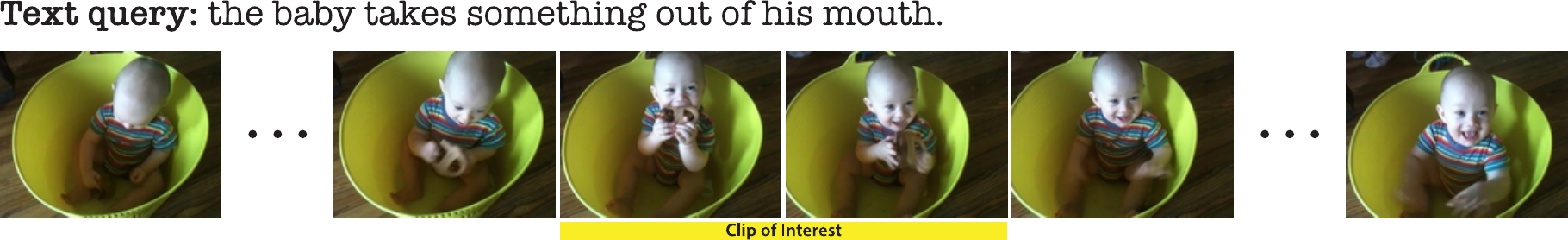}
\end{center}
\caption{An illustration of localizing clips of interest by natural language description.
    In this example, we are looking for the clip that the baby takes something out of his mouth. %MORE NEEDED?
The video clip above the yellow bar is the clip we are interested in.}
\label{fig:intro}
%\vspace{-0.2cm}
\end{figure*}

\section{Introduction}
% Our problem
%Video analysis has become a highlighted research topic in the area of multimedia and computer vision. It aims at understanding video content automatically, ranging from low-level visual and audio information to high-level multimedia semantics.
%Many challenging problems, \eg, video retrieval~\cite{lin2017hnip} and video event detection in videos~\cite{mazloom2016tagbook, song2018extracting}, have been widely studied in recent years.

Representation learning for videos is one of the fundamental problems in video analysis.
%Many approaches have been proposed to learn local representations~\cite{} and global representations~\cite{} for videos.
%the recent state-of-the-art methods usually follow the two steps.
To recognize if a concept occurs in a video, \ie, to classify video content, a typical way is encoding videos into video-level representation vectors. A classifier is then applied to the video-level vectors to determine the occurrence of pre-defined concepts.
%There are many Convolutional Neural Networks (ConvNets) derive frame-level or clip-level representations from raw video data by modeling spatial structures or spatio-temporal structures~\cite{simonyan2014two, tran2015learning, wang2016temporal}.
Frame-level or clip-level representations are extracted from raw video data by Convolutional Neural Networks (ConvNets) that models spatial structures or spatio-temporal structures~\cite{simonyan2014two, tran2015learning, wang2016temporal}. %, shi2017sequential, hou2018content}.
Video-level representations are generated by aggregating frame-level or clip-level local representations.
%{\color{red}{Frame-level or clip-level representations are learned from raw video data by Convolutional Neural Networks (ConvNets) that models spatial structures or spatio-temporal structures~\cite{simonyan2014two, tran2015learning, wang2016temporal}. }}

%%% To classify videos, usually aggregate local features (are generated by RGB/Flow/C3D ConvNets) into a global vector, and apply a classifier on this representation vector.
%%% Introduce RGB Flow C3D

%In video classification, there are usually two ways to learn a single global vector representation.
%One way is to generate frame-level or clip-level features, upon which an aggregating method is applied ~\cite{simonyan2014two,wang2016temporal}.
%%\cite{NIPS spatio-temporal}
%The other way is to model video spatio-temporal structures with 3D Convolutional Neural Networks (ConvNets) in an end-to-end way~\cite{tran2015learning}.

%%% Representation learning to Existing tasks

%%% To better understanding the details of videos, \eg, when did the concept occurs.

The video classification task only concerns about the occurrence of a target concept, \eg, an action or an object, in a video, but ignores in which frames the concept can be seen.
Someone may be more interested in temporally localizing the concept in untrimmed videos, \eg, the temporal action detection task~\cite{Lin:2017:SST:3123266.3123343, cdc_shou_cvpr17, Xu_2017_ICCV, Zhao_2017_ICCV}.
%Existing methods for temporal action detection applied more detailed analysis on the temporal domain, pushed video understanding into finer scale.
However, the existing action detection datasets have pre-defined actions with limited number of categories. For example, in THUMOS 14~\cite{THUMOS14} and ActivityNet~\cite{caba2015activitynet}, there are only 20 and 200 categories, respectively.
%In addition, the action categories are usually
It is difficult to adapt the trained models to unseen concept categories outside the training set,
which limits the usage of the detection models in real-world applications. 
%The models trained on the dataset can be difficult to generalized to other categories.
For example, to find a specific clip in a movie, instead of describing the target clip using a pre-defined concept, describing the target clip by a sentence is a more natural way.
%%%   WHY LIMIT IN REAL-WORLD APPLICATIONS

%Another line of recent works is action detection, which aim k
%In a untrimmed long video, to test the recognition ability of the model, it can be if the model generates a label prediction together with to test the temporal localization ability on localize actions
%Some have focus on video action detection, where the model need to detect pre-defined actions in a long video sequence,
%~\cite{} proposed to xxx.
%Recently, Hendricks~\etal~\cite{Hendricks_2017_ICCV} and Gao~\etal~\cite{Gao_2017_ICCV} introduced a new task, \ie, temporally localizing video clips by natural language descriptions, along with the new DiDeMo dataset and the Charades-STA dataset.
Recently, Hendricks~\etal~\cite{Hendricks_2017_ICCV} and Gao~\etal~\cite{Gao_2017_ICCV} introduced a new task, \ie, temporally localizing video clips by natural language descriptions. Hendricks~\etal~collected the new DiDeMo dataset, while Gao~\etal~utilized the videos from the Charades dataset~\cite{sigurdsson2016hollywood} and annotated the Charades-STA dataset.
%{\color{red}{Recently, Hendricks~\etal~\cite{Hendricks_2017_ICCV} and Gao~\etal~\cite{Gao_2017_ICCV} introduced a new task, \ie, temporally localizing video clips by natural language descriptions. Hendricks~\etal~collected the new DiDeMo dataset, while Gao~\etal~ utilized the videos from the Charades dataset and annotated the  Charades-STA dataset.}}
Natural language descriptions are more expressive than a pre-defined label described by a single word or a short phrase.
For example, in the description of ``the baby takes something out of his mouth'', there are different object categories (``baby'', ``mouth''), attributes (``his'') and human-object interactions (``take something out of'').
To localize the described clip in a video sequence, the visual model has to recognize objects, actions, spatial relations, and also the context of the event.
Standard action detection algorithms only localize actions at a coarse level. In this new task, the model has to understand both language descriptions and video content in finer scales.
These datasets contain
%The DiDeMo dataset has 
%thousands
a great number 
of descriptions, covering a large vocabulary and a variety of complex sentence structures.
With such rich information, though challenging, this task enables us to learn a model that generalizes to new natural language descriptions.

The existing methods for video clip localization by natural language description follow traditional cross-modal retrieval approaches~\cite{frome2013devise}, embed visual information and language information individually~\cite{Hendricks_2017_ICCV, Gao_2017_ICCV}.
Hendricks~\etal~\cite{Hendricks_2017_ICCV} used a ranking loss to train the embedding functions, so that the language feature is closer to the ground-truth clip than all negative clips.
Gao~\etal~\cite{Gao_2017_ICCV} merged clip visual embedding and language embedding by a multi-modal processing module, followed by alignment score estimation and location regression.
%The first and only existing method for video clip localization by natural language description embeds visual information and language information individually~\cite{Hendricks_2017_ICCV}.
%Hendricks~\etal~\cite{Hendricks_2017_ICCV} use a ranking loss to train the embedding functions so that the language feature is closer to the ground-truth clip than all negative clips.
%Both work are breakthroughs into the new area, but ignored detailed video-language multi-modal modeling.
%% Both methods ignored detailed 
%% To address these problems,
In this paper, we propose a novel attentive sequence to sequence translator (ASST) for this challenging task. Specifically, we make the following contributions.

First, we introduce a finely calibrated video-language attentive mechanism for understanding rich semantic descriptions.
A bi-directional Recurrent Neural Network (biRNN) is utilized to parse text descriptions in a forward and a backward way.
The method in \cite{Hendricks_2017_ICCV} generates a vector for each fixed-length clip by average pooling, during which it loses the local information of each individual frame.
The method in \cite{Gao_2017_ICCV} keeps temporal structures of sampled video clips by temporal convolutions, but ignores detailed relation with natural language descriptions.
Likewise, the methods in \cite{Hendricks_2017_ICCV} and \cite{Gao_2017_ICCV} only generate one vector for each sentence, without word-level information.
Our method works in a much finer manner and attends every meaningful word or phrase to each frame. For example, given a sentence of $6$ words and a video of $100$ frames,
%our attention is a matrix of $6 \times 100$.
we generate an attention weight matrix of size $6 \times 100$.
Richer information as represented by the attention weight matrix enables our method to perform a finer video-language matching.
Our attention mechanism aligns language descriptions and video content, which benefits the understanding of both semantic description and video content.
For example, when localizing ``we can see someone in a mask'', the target video clip will attend to the word ``mask'' as opposed to the entire sentence.  In this way, a more accurate video-language representation can be generated.

Second, we design a hierarchical network architecture
%{\color{red}{We didn't mention ``hierarhical network``'' in  later sections???}}
for jointly learning video-language representations with multiple granularities.
%unified video-description sequence representation for frame level vision-language matching.
The %dilated Convolutional Neural Network (ConvNet) 
%fully convolutional video network
hierarchical visual network
%encoder
models videos at multiple temporal scales to extract subtle details of video content as well as 
global context. 
Lower layers in the hierarchy exploit local video content, which are then integrated at a higher layer to model temporal dependencies in a longer duration. In other words, lower layers provide finer exploration while higher layers provide context information. 
For example, given a video of birthday party, we first model each frame with details, then we model three sub-events including  ``cut a birthday cake'', ``sing a song'', and ``eat the cake'' at a higher layer, and lastly we model the entire event of birthday party in the highest layer.
A robust joint video-language representation is then generated, which preserves fine-scale details, frame and clip dependencies as well as temporal context of video content.
%encoding temporal context from subtle details to global video-level context.

We evaluate our ASST on two large-scale datasets for localizing clips by natural language descriptions. On the DiDeMo dataset, it achieves $32.38\%$ in Rank$@1$ and outperforms the state-of-the-art by $4.28\%$.
On the Charades-STA dataset, we obtain $37.04\%$ in Rank$@1, IoU=0.5$. Notably, it outperforms the state-of-the-art
%by $15.62\%$
by a large margin of $15.62\%$.

\section{Related work}
%\noindent{\textbf{Object detection}}.
%$Object detection is one of the fundamental tasks in computer vision.
%$Many deep learning based approaches have been proposed for this problem~\cite{dai2016r, Huang_2017_CVPR, Lin_2017_ICCV, liu2016ssd, ren2015faster}.
%$Currently, the general idea of state-of-the-art object detection models is to build detection anchors on the feature layers of the backbone network.
%$In some recent works, the top-down module has been proven useful for global context awareness~\cite{lin2016feature, Lin_2017_ICCV, shrivastava2016beyond}.
%$Global context from the entire video sequence can be vital for video clip understanding.
%$We thus incorporate a similar top-down module to learn global context.

\noindent{\textbf{Action detection in videos}}.
In action detection, the goal is to localize interested video clips of given action labels.
This task is usually divided into two branches: temporal detection and spatio-temporal detection.
For temporal detection~\cite{THUMOS14}, the model has to output the start and end time of each action clip.
For spatio-temporal detection~\cite{pantofaru2017ava}, besides temporally localizing the actions, the model should also generate the spatial locations of the actions for each frame.
%the spatial locations of target actions on each frame are also needed.
We do not consider spatial localization in this paper.

% Temporal detection
Two-Stream ConvNets~\cite{simonyan2014two, wang2015towards} %, wang2017two} 
and 3D-ConvNets~\cite{tran2015learning} have been proposed to model video sequences for action recognition. These methods have been widely used in temporal action detection~\cite{Lin:2017:SST:3123266.3123343, cdc_shou_cvpr17, Xu_2017_ICCV, Zhao_2017_ICCV}
and other video analysis tasks, \eg, video captioning~\cite{venugopalan15iccv}.
%\cite{Kalogeiton_2017_ICCV2, Kalogeiton_2017_ICCV, Sigurdsson_2017_ICCV}. % MORE CITATION?
Shou~\etal~\cite{cdc_shou_cvpr17} applied frame-level classification via a modified 3D-ConvNet, and then generated temporal window prediction by merging frame-level prediction.
Lin~\etal~\cite{Lin:2017:SST:3123266.3123343} adopted Single Shot Detector~\cite{liu2016ssd} to detect interested actions in one shot.
Xu~\etal~\cite{Xu_2017_ICCV} tried to tackle the problem of temporal proposal generation. They incorporated an RoI-pooling layer, followed by a classification branch and a temporal coordinate regression branch.
Zhao~\etal~\cite{Zhao_2017_ICCV} achieved a good performance by hard coding context regions, which enables the detector to model temporal context. They also proposed a sophisticated region merging strategy.
%\red{(Please add one sentence talking about our method,)}
We aggregate contextual temporal structures via dilated convolutional layers, followed by squeezing-expansion layers.
Our model jointly encodes neighboring and all observable visual information, which is more flexible for temporal structure modeling.

\noindent{\textbf{Natural language object localization}}.
Natural language object localization~\cite{hu2016natural, mao2016generation} is to spatially localize objects of interest in images by natural language descriptions.
%This is a similar task to the natural language localization in videos.
Mao~\etal~\cite{mao2016generation} encoded each possible region by a Long Short-Term Memory (LSTM), and trained the model in a discriminative way by using a softmax loss over all encoded regions.
Hu~\etal~\cite{hu2016natural} ranked each object proposal by considering spatial configuration, local visual information and global visual information.
%Hu~\etal~\cite{hu2016natural} ranked each object proposal by scoring given natural language description conditioned by spatial configuration, local visual information and global visual information.
Rohrbach~\etal~\cite{rohrbach2016grounding} used attention mechanism to choose the region that could be best used for description reconstruction.
%Nagaraja~\etal~\cite{nagaraja2016modeling} localized referred objects by understanding referring expressions via integrating context between objects.
Liu~\etal~\cite{liu2017referring} extracted image attributes from image regions proposals, and measured the similarity between image region attribute embeddings and textual embeddings.
%Hu~\etal~\cite{hu2017modeling} parsed describing sentence in detail into subject, relationship and object. Then image proposed are scored by 
In this paper, we focus on natural language video clip localization. We propose a novel ASST to jointly learn a video-language representation.

%VIDEO-LANGUAGE ALIGNMENT?

%CLEVR?

%VQA?

%\noindent{\textbf{Sequence to sequence learning}}.
%Sequence to sequence learning is a important research area in recent years, and being widely applied in many areas, \eg, machine translation and captioning \etc.
%Most work applies Recurrent Neural Network (RNN) to build encoder for parsing input sequence and decoder for generating output sequence~\cite{donahue2015long, sutskever2014sequence}. 
%Recently, Gehring~\etal~\cite{gehring2017convolutional} adopt Convolutional Neural Network (CNN) with attention module to build a convolutional sequence to sequence learning model. On each generation step, this model generates one element by hierarchically attending convolved input sequence and causal convolved previously generated output sequence.

%\noindent{\textbf{Multi-modal representation learning}}.
%Multi-modal representation learning is a research area about exploring relevance cross data modality, like image and natural language. Common cross-modal solutions is mapping cross-modal data independently into a unified semantic space, followed by distance measuring~\cite{frome2013devise}. These models are usually optimized by ranking loss or triplet loss. Karpathy~\etal~\cite{karpathy2015deep} apply bi-directional RNN for language modeling and R-CNN for image region modeling.

\noindent{\textbf{Video clip localization by natural language descriptions}}.
%\red{(It is a new task. MCN is proposed, it is a ranking based method)}
Video clip localization by natural language descriptions is a new task introduced recently~\cite{Hendricks_2017_ICCV,Gao_2017_ICCV}.
%is a new task introduced by Hendricks~\etal~\cite{Hendricks_2017_ICCV}. 
Hendricks~\etal~\cite{Hendricks_2017_ICCV} proposed a Moment Context Network (MCN) for this task. MCN is a ranking based method. It encodes natural language description and video clip information individually, and ranks video clips by measuring distance between video clip embeddings and sentence embeddings. MCN considers representations from two modals separately, ignores detailed relevance between two modals.
Gao~\etal~\cite{Gao_2017_ICCV} fused video feature and language feature by a multi-modal processing module which consists of addition, multiplication and fully connected operations. This Cross-model Temporal Regression Localizer (CTRL) builds an alignment score estimator and a location regressor on the fused representation to produce clip prediction.
Our ASST explores cross-modal relevance at finer granularities, which is able to better exploit detailed information.

%CCA

%Classic multi-modal representation learning work for dealing with natural language and visual data are usually mapping them independently into a unified semantic space, Some recent work start to 

%Recurrent neural network based sequence to sequence learning~\cite{donahue2015long, sutskever2014sequence} and convolution based sequence to sequence learning~\cite{gehring2017convolutional} have shown good performances in many applications.  

%TODO

%Our model is a task dependent one-to-one video translation model. 

% In our network, we model our temporal video feature in a sequence to sequence learning style.

%Bijection
%one-to-one correspondence

\begin{figure*}[ht]
\begin{center}
\includegraphics[width=.95\linewidth]{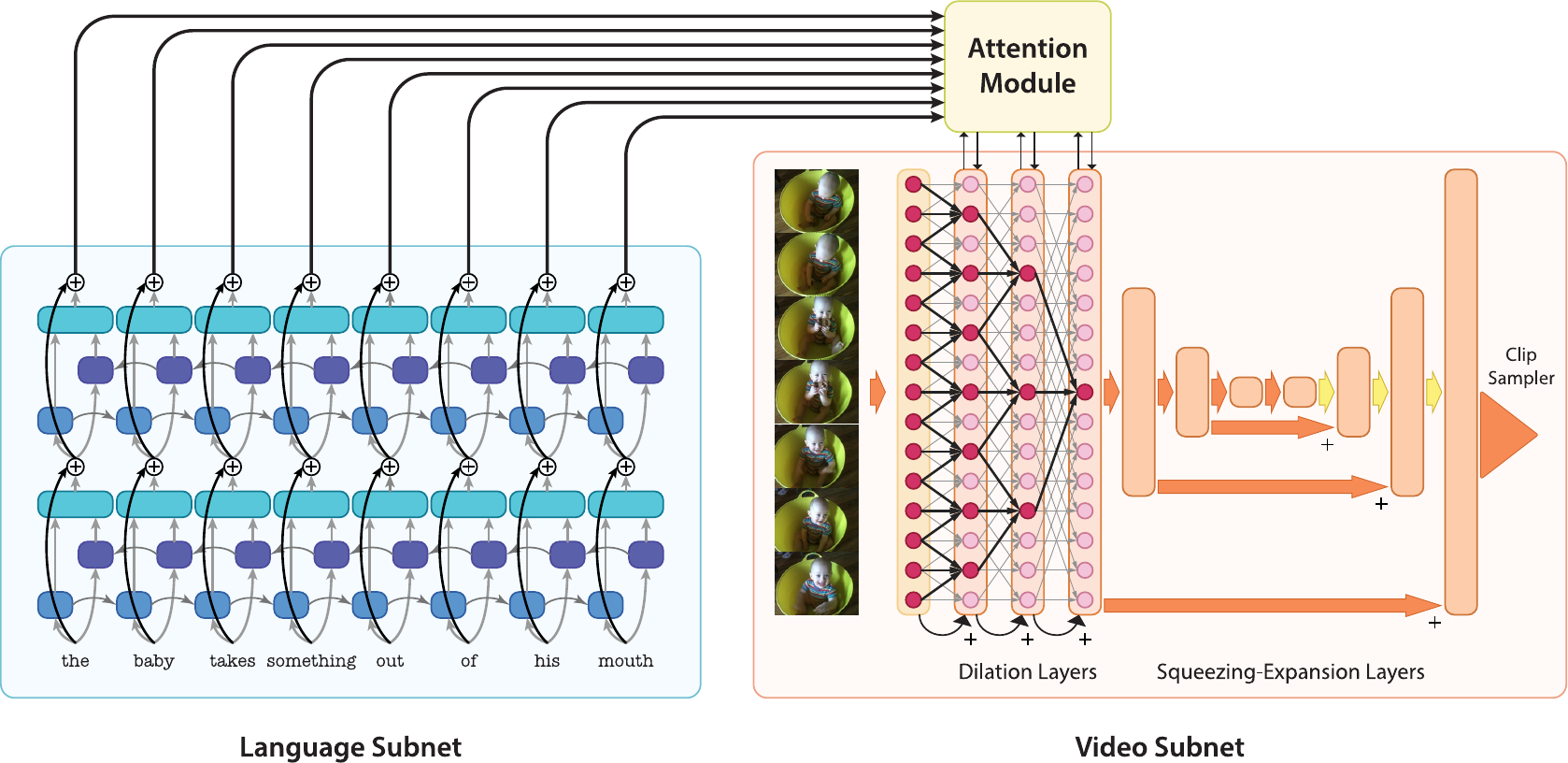}
\end{center}
\caption{An illustration of our model. Our model consists of an RNN-based language subnet (left) and a ConvNet-based video subnet (right).
    The two subnets are integrated via an attention module (top) on every dilation layer and squeezing layer.
    In this example, the language subnet has one stacked LSTM layer. For the video subnet, there are three dilation layers, followed by three squeezing layers and three expansion layers.
    %Orange arrows represent convolution operations.
%representation layers.
The output of the last dilation layer is used as the input of first squeezing layer. Best viewed in color.}
\label{fig:network}
%\vspace{-0.2cm}
\end{figure*}

\section{Our approach}

%Comparing to the classic detection task, 
%the target to be detected, \ie, 
%natural language descriptions, have much more varieties than class labels. They are also not enumerable, which makes classic detection approaches unfeasible for our task.
In our work, we leverage natural language information as a flexible network module to translate visual information. We exploit detailed connections between natural language information and visual information on multiple hierarchies as well. We directly sample possible clips from the final video-language joint representation sequence generated by our ASST for clip localization.

%\section{Attentive Sequence to Sequence Translation}
As illustrated in Figure~\ref{fig:network},
the proposed attentive sequence to sequence translator (ASST) consists of 
three
%two?
components: a video subnet for modeling video content, a language subnet for modeling natural language description and a multi-stage cross-modal attention module.
%Our ASST is end-to-end trainable.
%These two subnets are combined into one end-to-end model by a multi-stage attention module.

% TODO: The purpose of doing so

\subsection{Video subnet}
Our video subnet is shown on the right part in Figure~\ref{fig:network}.
The video subnet first translates the input visual feature sequence into a representation sequence.
The length of the representation sequence and the length of the input feature sequence are identical.
Each element in this final representation sequence contains temporal visual contextual information, ranging from subtle details to global context, as well as clues from natural language descriptions.
Then, video clip representations are obtained by directly sampling clips on the final representation sequence.
%The base network consists of context layers and representation layers.
%On the last stage,
%we build different detection anchor groups on the final representation sequence for different tasks. 
%we directly sample clips on the final representation sequence depend on the task we are handling.
%We demonstrate our video subnet from these two aspects.

\subsubsection{Visual feature learning}
The video subnet is a 1D %fully-
convolutional network spanned on temporal domain.
We take a sequence of vectors $\{\mathbf{x}_{0}, \mathbf{x}_{1}, ..., \mathbf{x}_{m-1}\}$ ($\{\mathbf{x}\}$ for simplicity) with dimension $d_v$ as the input of the video subnet,
where $m$ is the number of frame-level features in the input sequence.
%%sequence length of input visual feature.
We also denote the temporal length in seconds of the input video sequence
%%spanned on temporal domain
as $\tau$.
%%\red{There are two process, One is xxx, the other is xxx. works together xxx}
%%There are two processes.
%%The subnet
Visual feature modeling
%learning
consists of two stages.
One is
%attentive
local temporal feature modeling via stacked dilated convolution. The other one is global temporal context modeling via a squeezing-expansion process.
The video subnet transforms an input sequence 
%$\{\mathbf{x}_{0}, \mathbf{x}_{1}, ..., \mathbf{x}_{t-1}\}$
$\{\mathbf{x}\}$
to a representation sequence with same length $\{\mathbf{x}'_{0}, \mathbf{x}'_{1}, ..., \mathbf{x}'_{m-1}\}$ ($\{\mathbf{x'}\}$ for simplicity) via these two stages.
%with a same length as input sequence ${\mathbf{x}}$.

\noindent\textbf{Dilation Layers}. We first model the frame-level input features with their neighboring frames.
This process
builds
%learns
temporal hierarchical structure of the input sequence via stacked dilated convolution layers~\cite{kalchbrenner2016neural, Lea_2017_CVPR, oord2016wavenet, yu2015multi}.
%(\red{The context layer xxxxx, illustrate the usage.})
%We first feed the input visual feature sequence into 
%It is done by several dilated convolution layers~\cite{kalchbrenner2016neural, Lea_2017_CVPR, oord2016wavenet, yu2015multi} for temporal context learning.
The dilation rates are doubled from each layer forward.
Stacked dilated convolution increases receptive field in an exponential speed as the network goes deeper, while the sizes of feature maps remain unchanged.
These layers model temporal contextual structures at multiple time scales for every input frame.
%Natural language descriptions interact with visual information within context layers.
%We will demonstrate this part later. 
In Figure~\ref{fig:network}, we show a model with three dilation layers.
We additionally apply shortcut connections~\cite{he2016deep} between every two adjacent layers.
%by adding feature map before last dilated convolution to the next corresponding feature map.
Shortcut connections shorten the information path from the bottom to the top of the network, which makes the training process easier.
%\red{(how the shortcut connection works. directly pass the output to , the apply add. retain the information from bottom layer)}
%The shortcut connections are very helpful for model training.
After the above operations, every element in the last dilation layer perceives a large temporal window.
%and natural language information.
%However, context layers only models local contextual temporal structure of input sequence without global visual information, 
%there is still no global context from the entire video sequence, 
%while global context is vital for localizing clips in entire input video.

%linear? bilinear? interpolation
\noindent\textbf{Squeezing-Expansion Layers}. %\red{(Representation layers encode global context. Please illustrate the benefit, why need this)}
%\textbf{Residual}
The dilation layers demonstrated above contain only local temporal contextual structures.
The elements in the final dilation layer do not observe the entire input visual sequence.
However, the understanding of global context from the entire 
video
%clip
is crucial for localizing clips in videos.
For example, someone may be interested in localizing the clip when a certain target event happens the second time.
To distinguish the positive clip from visually similar negative clips, the model has to understand the global context to make accurate predictions.
%what is current observed local visual information representing under the context of entire video.
To model global context information, we generate our final representation sequence through a squeezing phase followed by an expansion phase (Figure~\ref{fig:network}).
The output of the last dilation layer is used as the input of the first squeezing layer.
%%The architecture of ``Representation Layers'' is similar to top-down module in object detection~\cite{lin2016feature, Lin_2017_ICCV, shrivastava2016beyond}. 
%The architecture of ``Squeezing-Expansion Layers'' can be viewed as a 1D version of top-down module for object detection~\cite{lin2016feature, Lin_2017_ICCV, shrivastava2016beyond}.
The architecture of ``Squeezing-Expansion Layers'' is inspired by top-down module for object detection~\cite{lin2016feature, Lin_2017_ICCV, shrivastava2016beyond}. %We made several modifications for the clip localization task. 
%%Though representation layers are built on context layers, there is no global representation in this process. %???
The model further encodes global context information into the final representation sequence through this squeezing-expansion process.

The squeezing phase consists of several convolution layers with stride $2$ and kernel size $3$, and generates a global representation vector. This vector summarizes visual information and temporal structures across all the input visual features.
It will be used as global context background for further representation
%learning.
modeling.
%\red{more details}

%\red{The expand phase expand the representation in a reverse order}
The expansion phase and the squeezing phase are connected by a convolution operation.
During the expansion phase, we expand the global representation vector in a reverse order as we did during the squeezing phase. Expansion stops when it reaches the size of the last dilation layer.
%Then we expand the global representation to each squeezed layer in a reverse order.
We apply a convolution with kernel size $1$ on each feature map from the squeezing phase, while the output is added to the expanded feature map with the same size.
The result feature map includes both global contextual information (from expanded feature maps) and local contextual information (from convolved feature maps in the squeezing phase).
We use linear interpolation as our expansion operation. 
The output of the last layer is our final representation sequence.

With dilation layers and squeezing-expansion layers, our translator is able to transform input visual sequence into a representation sequence, exploiting both global and local temporal contextual structures.

%\red{Another paragraph, with the context layer and the representation layer, our translator is able to xx, the representation xxx}

%a hierarchical structure for the network to jointly model language description and video content.

%During representation layers, the input features are first gradually squeezed into one global representation vector, then expanded to the size of each previous representation layers in a reverse order. Every representation layer in squeeze phase is passed through a convolution convolution layer with kernel size $1$ and added to corresponding layer in expansion phase. We use linear interpolation as our expansion operation. The last layer is our final representation sequence.

%\subsubsection{Detection anchor building}
\subsubsection{Clip sampling}
% CHANGE NAME TO ``Clip sampling and Loss function building''
% CHANGE NAME TO ``clip sampler'' ?
% bilinear? linear? interpolation
% Possible regions / proposals are provided by dataset !!
We sample our 
%detection anchors
%clip samplers
target clips
on the final representation sequence $\{\mathbf{x}'\}$ by RoI-pooling~\cite{girshick2015fast} in temporal domain with linear interpolation.
For each clip sampler, we sample $7$ elements from the final representation sequence $\{\mathbf{x}'\}$ as the input.
%detection anchor

% IS THIS 'DILATED' NECESSARY?
%The detection anchor
The clip sampler is a submodule %mini module
consists of two stacked convolution layers with kernel size $3$.
%, which means the input for each detection anchor is $7$
%All detection anchors
All clip samplers
share weights.
Under different circumstances,
clip samplers can be built differently.
%in different circumstances.
We show the following two cases.

%We use different
%clip sampling
%anchor building
%strategies for different tasks.

%% Revise

%If there are pre-defined temporal segments, 
\noindent{\textbf{If there are pre-defined temporal segments}},
%For the natural language localization task, 
%Under this circumstance, 
we address this problem as a classification problem by enumerating pre-defined segments. We build one
%anchor
clip sampler
for each possible segment. Each
%anchor
sampler generates one scalar, which represents the confidence score for the corresponding video clip.
% SCORE?
%We also feed TEF (temporal endpoint features) into
%anchors
%clip samplers by concatenating TEF to every sampled vector
%detection anchors
%following \cite{Hendricks_2017_ICCV}.
Finally, a softmax loss function is applied over all
clip samplers
%anchors
to train our ASST in a discriminative
%way
manner
following \cite{mao2016generation}.
We denote this model as \textit{classification model}.

% FIGURE FOR ANCHOR BUILDING?

%If there are no pre-defined temporal segments
\noindent{\textbf{If there are no pre-defined temporal segments}}, 
%Under this circumstance, 
our model will have to generate clip proposals.
%In this circumstance, 
%For the action detection task, 
We build
%detection anchors
clip samplers similar to \cite{ren2015faster}.
%Except from natural language clip detection task, we can also build anchors for action detection.
In our model,
we have six groups of
%anchors. 
samplers.
%The
% DISTINGUISH LENGTH AND TEMPORAL LENGTH
The $i$-th clip sampler group consists of clips with length $l_{i}=\frac{\tau}{2^{i}}$, where $0 \leq i \leq 5, i \in \mathbb{N}$, and $\tau$ is the temporal length of the final representation sequence $\{\mathbf{x}'\}$ as we mentioned before.
We sample clips densely by placing every two adjacent samplers with distance $\frac{1}{3}l_{i}$. Therefore, there are $2^{i+2}-3$ samplers in the $i$-th group.
%These detection anchors share weights as well.
For each clip sampler, we predict a tuple
$(\hat{y}, \hat{d_c}, \hat{d_l})$, where $\hat{y}$ is confidence score of the presence of target video clip, $\hat{d_c}$ represents clip center deviation and $\hat{d_l}$ is the $\log$ difference from the sampled clip length.
%with class label $y$, center deviation $d_c$ and $\log$ difference from the default clip length $d_l$.
During inference, for each sampled clip with temporal center coordinate $c_s$ and clip length $l_s$,
the predicted temporal window is calculated by,
%the temporal location of each clip sampler's prediction is calculated by,
%anchor
%of length 
%anchor
\begin{equation}
(\hat{c}, \hat{l}) = (c_{s}+\hat{d_{c}}l_s, e^{\hat{d_l}}l_s),
\label{eq:boxPred}
\end{equation}
%where $c_0$ is the center temporal coordinate of the default window,
%$l_0$ is the length of each clip,
where $\hat{c}$ is the center coordinate of the predicted window and $\hat{l}$ is the length of the window.
%We use standard $(n_c+1)$-class softmax loss, where $n_c$ is the number of classes.
We use $2$-class softmax loss to train the confidence score of presence $\hat{y}$.
% determine whether this clip is positive.
Smooth-$L1$ loss is used to train temporal coordinate regression factors $d_c$ and $d_l$ following most object detection literatures~\cite{dai2016r, Huang_2017_CVPR, Lin_2017_ICCV, liu2016ssd, ren2015faster}.
%\red{cite more papers}
We denote this model as \textit{detection model}.

\subsection{Language subnet}
%\subsubsection{Simple language}
%\subsubsection{RNN context}
The architecture of our language subnet is shown on the left part in Figure~\ref{fig:network}.
We use pre-trained word embedding sequence $\{\mathbf{w}_0^0, \mathbf{w}_1^0, ..., \mathbf{w}_{n-1}^0\}$
%($\{\mathbf{w}^0\}$ for short)
as linguistic inputs.
Stacked bi-directional LSTM layers are built on top of word embedding sequence for sentence context modeling~\cite{wu2016google}. We also add shortcut connections from the input to the output of each LSTM layer for more efficient training.
%In practice, we use only one layer of bi-directional LSTM.

% REMOVE ???
%We modify the standard bi-directional LSTM by adding one convolution operation with kernel size $1$ after concatenating forward and backward output states. {\color{red}{language}} It is helpful for bi-directional LSTM's residual learning. We also apply a convolution operation with kernel size $1$ on the input word embeddings to adjust dimension for residual learning.

\begin{equation}
\begin{aligned}
\{\mathbf{w}_0^{i}, \mathbf{w}_1^{i}, ..., \mathbf{w}_{n-1}^{i}\} = &~ biRNN_i(\{\mathbf{w}_0^{i-1}, \mathbf{w}_1^{i-1}, ..., \mathbf{w}_{n-1}^{i-1}\})\\
+ &~ \{\mathbf{w}_0^{i-1}, \mathbf{w}_1^{i-1}, ..., \mathbf{w}_{n-1}^{i-1}\},
\end{aligned}
\label{eq:birnn}
\end{equation}
where
%$n$ is the number of input words,
$\{\mathbf{w}^i\}$ is the result word-level feature sequence of the $i$-th layer.
%\red{to add $w_n^{i}$}
After 
%We stack 
$p$ layers of bi-directional LSTMs,
%in our experiments.
the output sequence $\{\mathbf{w}_0^{p}, \mathbf{w}_1^{p}, ..., \mathbf{w}_{n-1}^{p}\}$ ($\{\mathbf{w}^p\}$ for simplicity) is the output of the language subnet which will be used as the input for the cross-modal attention mechanism.
%\red{CONV}

%GLU!!!
We choose Gated Linear Unit (GLU)~\cite{dauphin2016language} as the non-linearity function of convolutions for language modeling. GLU processes a input vector $[\mathbf{q}_0, \mathbf{q}_1]$ by,
\begin{equation}
GLU([\mathbf{q}_0, \mathbf{q}_1]) = \mathbf{q}_0 \odot \sigma(\mathbf{q}_1),
\end{equation}
where $\mathbf{q}_0$ and $\mathbf{q}_1$ are the inputs to the GLU with dimension $d$, $\sigma$ is the sigmoid function, and $\odot$ is element-wise multiplication.
GLU has been used in context modeling in natural language processing~\cite{dauphin2016language,gehring2017convolutional}. The gate $\sigma(\mathbf{q}_1)$ dynamically controls data flow via the gating mechanism.
GLU shows superior performance than ReLU in our preliminary experiments.
%\red{GLU has been used in xxx, how does it work, what's the benefit.}
% GLU is better because of experiments?
% Conv?

\subsection{Cross-modal attention module}

\begin{figure}[t]
\begin{center}
\includegraphics[width=.95\columnwidth]{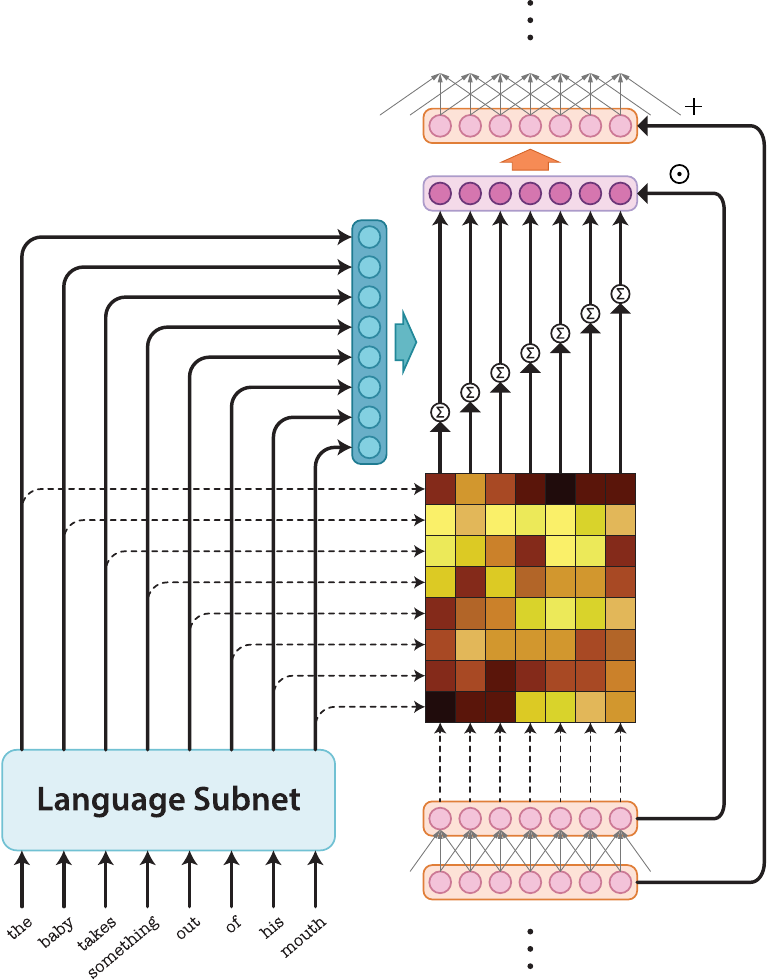}
\end{center}
\caption{An illustration of the attention module for combining the video subnet and the language subnet on one dilation layer.
    This module takes language feature (left) and video feature (bottom right) as inputs.
    It first generates an attention weight matrix from both input features, and then attends language feature for every video frame. This attended language feature multiplies with input video feature.
%works as a gate
%.
%Dynamic filter
After a convolution operation, shortcut connection is used to connect visual feature maps across adjacent % dilated
layers.    
%    cross dilated layers. \red{cross??}
Best viewed in color.}
\label{fig:Attention}
\end{figure}

To combine our video subnet and language subnet, we introduce a cross-modal attention module that combines visual and linguistic information.
%similar to the attention module proposed by Gehring~\etal~\cite{gehring2017convolutional}. 
The attention module attends word-level representations for each frame, discovers connection between word-level linguistic information and frame-level visual information.
It yields more detailed understanding of video content and description semantics.
%Our attention module 
Comparing to using sentence-level linguistic feature, using word-level linguistic feature also shortens the path of gradient flow back to each word.
In addition, this attention module is applied on every dilation layer and squeezing layer.
Since different
%dilation
layers aggregate temporal contextual information at
%different granularities
different temporal scales, feeding language feature on every layer enables our model to exploit video-language relevance at multiple temporal scales.
%\red{rewrite}
Through this attentive cross-modal fusion process, our ASST translates the input visual sequence into the final representation sequence that consists of joint language-video representation, which can be easily recognized by clip samplers.
%\red{rewrite}
In Figure~\ref{fig:Attention}, we show our attention module for one dilation layer. 

For each visual layer after temporal convolution $\{\mathbf{v}_0, \mathbf{v}_1, ..., \mathbf{v}_{m-1}\}$ ($\{\mathbf{v}\}$ for simplicity),
the attention module takes language feature $\{\mathbf{w}^p\}$ and $\{\mathbf{v}\}$
%$\{\mathbf{w}_0, \mathbf{w}_1, ..., \mathbf{w}_{n-1}\}$ and visual feature $\{\mathbf{v}_0, \mathbf{v}_1, ..., \mathbf{v}_{m-1}\}$
as inputs.
%conv with kernel size 1?
We split language feature
$\{\mathbf{w}^p\}$
%$\{\mathbf{w}_0, \mathbf{w}_1, ..., \mathbf{w}_{n-1}\}$
into two equal-length sequences by convolutions with kernel size $1$:
$\{\mathbf{w}_0^a, \mathbf{w}_1^a, ..., \mathbf{w}_{n-1}^a\}$ 
for attention weight matrix computation
and $\{\mathbf{w}_0^v, \mathbf{w}_1^v, ..., \mathbf{w}_{n-1}^v\}$
for further feature computation.
%$\{\mathbf{w}^a\}$ is used for attention weight matrix computation. And $\{\mathbf{w}^v\}$ is used for further feature computation.
%\red{rewrite, how to get a and v}
We also derive $\{\mathbf{v}_0^a, \mathbf{v}_1^a, ..., \mathbf{v}_{m-1}^a\}$ from $\{\mathbf{v}\}$ %.
%$\{\mathbf{v}^a\}$ is used as another input
for attention weight matrix computation. The attention weight matrix is computed as,
%\red{rewrite, v}
% \alpha or A ?
\begin{equation}
\begin{aligned}
A'_{ij} &= \frac{\mathbf{w}_i^a \cdot \mathbf{v}_j^a}{d_a},  \\
A_{ij} &= \frac{e^{A'_{ij}}}{\sum_i e^{A'_{ij}}},
\end{aligned}
\end{equation}
where $d_a$ is the dimension of $\mathbf{w}_i^a$ and $\mathbf{v}_j^a$, and $A$ is the attention weight matrix with size $n \times m$. We use this matrix to compute vision
%dependent
attended language feature,
\begin{equation}
\mathbf{u}'_j = \sum_i A_{ij}\mathbf{w}_i^v.
\end{equation}
We then apply element-wise multiplication on $\{\mathbf{u}'\}$ and $\{\mathbf{v}\}$,
\begin{equation}
\mathbf{u}_j = \mathbf{u}'_j \odot \mathbf{v}_j.
\end{equation}

%$\{u\}'$ can be viewed as a generalized conditional normalization.
%$\{\mathbf{u}\}'$ can be viewed as a gate for visual context information from dilated convolution.

%CONDITIONAL NORMALIZATION?
%GATE
%DYNAMIC FILTER

$\{\mathbf{u}\}$ is our language augmented context for this %dilation
convolution layer.
We apply a BatchNorm~\cite{ioffe2015batch} followed by a ReLU to normalize $\{\mathbf{u}\}$.
%, same as other convolution layers.
%stabilize the model
Finally, 
we add normalized $\{\mathbf{u}\}$ to the feature map before the last temporal
%we convolve $\{\mathbf{u}\}$ once and add it with the feature map before the last
%dilated
convolution, as the input for the next
%dilated
%convolution
temporal convolution.

\section{Experiments}
\subsection{Datasets}
%We use DiDeMo~\cite{Hendricks_2017_ICCV} dataset for natural language video clip localization task. To verify the effectiveness of our standalone video subnet, We use THUMOS 14~\cite{THUMOS14} as our benchmark dataset for action detection task.
We use the DiDeMo dataset~\cite{Hendricks_2017_ICCV} and the Charades-STA dataset~\cite{Gao_2017_ICCV, sigurdsson2016hollywood} to evaluate our model for the natural language video clip localization task.
%To verify the effectiveness of our standalone video subnet, We use THUMOS 14~\cite{THUMOS14} as our benchmark dataset for action detection task.

\begin{comment}
\noindent{\textbf{THUMOS 14}}.
THUMOS 14 is an action recognition and detection dataset. There are $101$ classes for action recognition task and $20$ classes for detection task. For action detection task, usually untrimmed videos from validation set and testing set are used as training data and testing data. There are $1,010$ videos in the validation set and $1,574$ in the testing set respectively. Only $200$ and $213$ videos in these two sets contain annotations within the $20$ action classes. Most recent literatures use these $413$ videos only for training and evaluation.
%We use both settings in our experiments. \ie, $1,010$ training videos with $1,574$ testing videos and $200$ training videos with $213$ testing videos.
In our experiments, we use these $413$ videos following most recent literatures.
The metric for THUMOS 14 dataset action detection task is $mAP$ (mean average precision) on different $IoU$s (Intersection over Union).
%The IoU thresholds are $0.1$, $0.2$, $0.3$, $0.4$ and $0.5$.
$mAP@IoU~0.5$ is usually used in recent papers.
\end{comment}

\noindent{\textbf{DiDeMo}}.
DiDeMo is a dataset for natural language description localization in open-world videos. There are $33,005$, $4,180$ and $4,021$ video-description pairs in the training, validation and testing subsets, respectively. 
To annotate video clips, videos in DiDeMo are segmented every $5$ seconds. The max available video length in this dataset is $30$ seconds. The ground truth of each video-description pair is annotated by multiple people on these segments. Therefore, for each video, there are $21$ possible clips.
%Since annotations divide videos every $5$ seconds and max available video length is $30$ seconds, there are $21$ possible clips.
As we described previously, we address this task as a $21$-way classification task.
Following \cite{Hendricks_2017_ICCV}, we use Rank$@1$, Rank$@5$ and mean Intersection over Union ($mIoU$) as evaluation metrics.
% MORE NEEDED?

\noindent{\textbf{Charades-STA}}.
Charades-STA is a dataset for natural language description localization in indoor videos. There are $12,408$ and $3,720$ video-description pairs in the training and testing subsets, respectively\footnote{The authors did some cleaning to the dataset. Updated dataset and results can be found at \href{https://github.com/jiyanggao/TALL}{https://github.com/jiyanggao/TALL}}. % The author cleaned the dataset
There are no pre-defined segments provided as in the DiDeMo dataset. We consider this task as a $2$-way temporal detection task.
Following \cite{Gao_2017_ICCV}, we use Rank$@1,IoU=0.5$, Rank$@1,IoU=0.7$, Rank$@5,IoU=0.5$ and Rank$@5,IoU=0.7$ as evaluation metrics.

\subsection{Implementation Details}
We implement our ASST by TensorFlow~\cite{abadi2016tensorflow}.
The code is publicly available at \href{https://github.com/NeonKrypton/ASST}{https://github.com/NeonKrypton/ASST}.
%We will make our code publicly available upon acceptance.

%During training time, we use Stochastic Gradient Descent (SGD) with Momentum $0.9$ as our optimizer. 
% ADAM ???
In our experiments, we build our ASST with one bi-directional LSTM layer for language subnet, four dilation layers, six squeezing layers and six expansion layers for video subnet. Our language subnet takes GloVe word embeddings~\cite{pennington2014glove} as input.
Following previous literatures~\cite{Gao_2017_ICCV,Hendricks_2017_ICCV}, we use pre-trained Two-Stream ConvNets~\cite{simonyan2014two} feature and 3D-ConvNets~\cite{tran2015learning} (C3D) feature as the visual input.
Two-stream ConvNets consist of RGB network and optical flow network, which take original RGB images and optical flows as inputs, recognize static objects and motions in videos respectively. C3D takes a sequence of consecutive RGB frames as input. It models static objects and motions simultaneously by applying hierarchical 3D convolution operations.
The channel size of LSTM states, dilation layers and squeezing-expansion layers are $512$, $1,024$ and $512$.
We apply dropout with rate $0.5$ on input visual feature, and $0.8$ for the rest of the model.
%on input visual feature with rate $0.5$, dilation layers and LSTMs with rate $0.8$.
We use Adam~\cite{kingma2014adam} as our optimizer. The learning rate starts from $5\times 10^{-4}$, and multiplies $0.9$ every $2,500$ steps. The batch size we used for training is $128$.
We use different training strategies for two datasets. The details are as follows.

\begin{comment}	
For action detection task, we use Inception-BN trained on ImageNet training set as our RGB network, and Inception-BN trained on UCF101~\cite{UCF101} split $1$ training set~\cite{wang2016temporal} as our optical flow network. We apply range dropout with rate $0.4$ on input feature.
We use mOHEM to select targets to be optimized in each iteration.
%Online Hard Example Mining (OHEM)~\cite{shrivastava2016training} to select training targets in each batch.
Any sampled clip with $IoU$ greater than $0.5$ with any annotation is considered as a positive example.
%The training targets are selected by mOHEM.
We use Stochastic Gradient Descent (SGD) with momentum $0.9$ as our optimizer. The learning rate starts from $5\times 10^{-3}$ and multiplies $0.9$ every $1,000$ steps. The model samples $5$ frames per second. At training time, we randomly stretch or compress our model by a scale within $[0.8, 1.25]$, then sample a clip of random time from training videos.
During inference, we slide our model along every testing video.
Non Maximum Suppression (NMS) with threshold $0.3$ within same class is applied as post processing.
% STEP 9?
\end{comment}

%For DiDeMo dataset, 
\noindent{\textbf{DiDeMo}}.
Following \cite{Hendricks_2017_ICCV}, we use VGG16~\cite{simonyan2014very} trained on ImageNet~\cite{russakovsky2015imagenet} training set as our RGB network.
Inception-BN~\cite{ioffe2015batch} trained on UCF101~\cite{UCF101} split $1$ training set~\cite{wang2016temporal} is used as our optical flow network.
The frame sample rate is $\frac{128}{30}$, so that our ASST can observe entire $30$-second video in one observation.
% We randomly move our network by random time deviation within $[-0.2, 0.2]$ seconds as data augmentation. % Optional
There are more than one annotation for each video-description pair. We first filter out annotations which has no overlap with other annotations from the same pair. Then, we randomly choose one annotation from remaining annotations for each iteration. To stabilize predictions, we average predictions from multiple model checkpoints.

%For Charades-STA dataset, 
\noindent{\textbf{Charades-STA}}.
Following \cite{Gao_2017_ICCV}, we use C3D~\cite{tran2015learning} trained on Sports-1M~\cite{karpathy2014large} as our visual input.
The frame sample rate is $4$. 
During training, we randomly stretch or compress our model by a scale within $[0.8, 1.25]$, then sample a clip of random time from training videos.
Any clip sampler has an overlap $IoU \geq 0.5$ with ground truth clip is considered as a positive training sample. The positive-negative sampling ratio is set to $1:1$.
During inference, we slide our model along testing videos.
Non-Maximum Suppression (NMS) with threshold $0.8$ is applied during post processing.

%We apply dropout on input visual feature with rate $[0.5, 1]$, context layers and LSTM with rate $[0.8, 1]$.

%The frame sample rate are adjusted automatically.

%\subsection{Experimental results}
%\subsection{Experiments on DiDeMo}
\subsection{Comparison with other methods}

\begin{table}[t]
\footnotesize
\begin{center}
\begin{tabular}{|l|c|c|c|}
\hline
Method & Rank$@1$ & Rank$@5$ & $mIoU$ \\
\hline\hline
Frequency Prior~\cite{Hendricks_2017_ICCV} & $19.40$ & $66.38$ & $26.65$ \\
CCA~\cite{Hendricks_2017_ICCV} & $18.11$ & $52.11$ & $37.82$ \\
MCN~\cite{Hendricks_2017_ICCV} & $28.10$ & $78.21$ & $41.08$ \\
\hline
%Ours RGB & $25.64$ & $69.24$ & $39.30$ \\
%Ours Flow & $30.37$ & $73.54$ & $45.64$ \\
%Ours Two-stream & $30.64$ & $75.75$ & $45.95$ \\
Ours & $\mathbf{32.38}$ & $\mathbf{78.44}$ & $\mathbf{47.49}$ \\
\hline
\end{tabular}
\end{center}
\caption{Natural language localization in videos results on test subset of the DiDeMo dataset.
%Our ASST outperforms all baselines on all metrics.
}
\label{tab:DiDeMo_Res}
\end{table}

\begin{table}[t]
\footnotesize
\begin{center}
\begin{tabular}{|l|c|c|c|c|}
\hline
Method
& \begin{tabular}{@{}c@{}} R$@1$ \\ $IoU$=$0.5$\end{tabular}
& \begin{tabular}{@{}c@{}} R$@1$ \\ $IoU$=$0.7$\end{tabular}
& \begin{tabular}{@{}c@{}} R$@5$ \\ $IoU$=$0.5$\end{tabular}
& \begin{tabular}{@{}c@{}} R$@5$ \\ $IoU$=$0.7$\end{tabular} \\
\hline\hline
%VSA-RNN~\cite{Gao_2017_ICCV} & $10.50$ & $4.32$ & $48.43$ & $20.21$ \\
%VSA-STV~\cite{Gao_2017_ICCV} & $16.91$ & $5.81$ & $53.89$ & $23.58$ \\
%CTRL~\cite{Gao_2017_ICCV} & $23.63$ & $8.89$ & $58.92$ & $29.52$ \\
CTRL~\cite{Gao_2017_ICCV} & $21.42$ & $7.15$ & $59.11$ & $26.91$ \\
%21.42	7.15	59.11	26.91
\hline
Ours & $\mathbf{37.04}$ & $\mathbf{18.04}$ & $\mathbf{68.12}$ & $\mathbf{38.28}$ \\
\hline
\hline
\begin{tabular}{@{}l@{}} Ours w/\\Two-Stream\end{tabular}
%& $\mathbf{49.60}$ & $\mathbf{29.22}$ & $\mathbf{75.78}$ & $\mathbf{48.58}$ \\
& $\mathbf{42.72}$ & $\mathbf{24.06}$ & $\mathbf{71.32}$ & $\mathbf{43.98}$ \\

\hline
\end{tabular}
\end{center}
\caption{Natural language localization in videos results on test subset of the Charades-STA dataset.}
\label{tab:Charades_Res}
\end{table}

%Successes

%Failures
\begin{figure*}[t]
\begin{center}
\includegraphics[width=.95\linewidth]{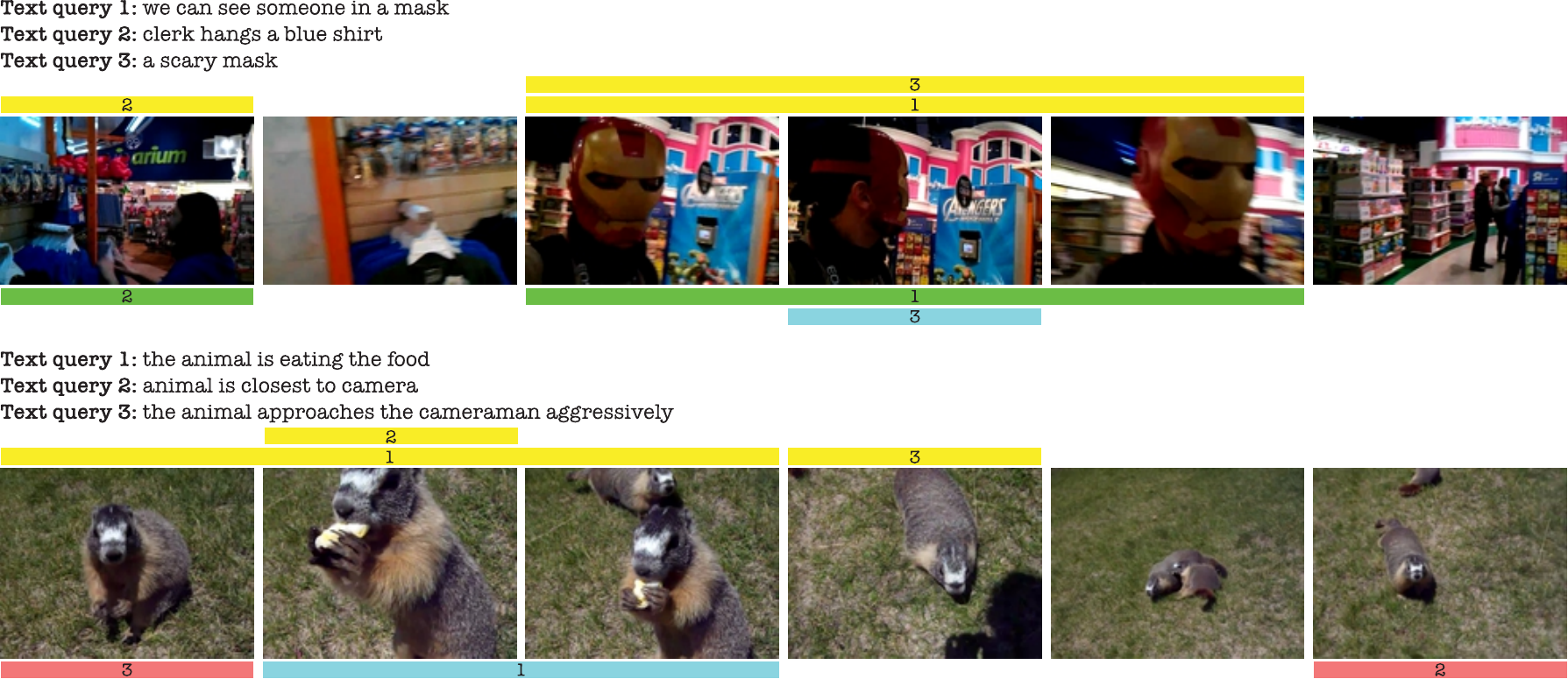}
\end{center}
\caption{Visualization of our ASST's experimental results of Rank$@1$. First two examples are from the DiDeMo dataset, and the last two are from the Charades-STA dataset.
The yellow bars above videos indicate ground truths. The bars under videos indicate our predictions. A green bar represents our prediction is correct ($IoU \geq 0.7$ for Charades-STA). A blue bar represents our prediction overlaps with ground-truth annotation, but not accurate ($0.5 \leq IoU < 0.7$ for Charades-STA). A red bar represents our prediction does not overlap with ground truth ($IoU < 0.5$ for Charades-STA). Best viewed in color.}
\label{fig:result}
\end{figure*}

%%\subsubsection{Natural language description localization}
%\subsubsection{Comparison with other models}

\noindent{\textbf{DiDeMo}}.
We perform experiments on the DiDeMo dataset for natural language description localization in videos. Table~\ref{tab:DiDeMo_Res} shows our results and the baselines provided in \cite{Hendricks_2017_ICCV}.

Our model outperforms MCN in all three metrics.
In Rank$@5$, our ASST is better than MCN by $0.23\%$.
In Rank$@1$ and $mIoU$, our ASST significantly outperforms MCN by $4.28\%$ and $6.41\%$, respectively.
In Figure~\ref{fig:result}, we show some of our localization results.
In the first example, our ASST successfully localized the sentence ``a girl with a blue shirt and black backpack speaks into the camera'' and ``there is a mountain''.
For the third text query, ``woman sharing a landscape view'', our ASST failed to generate an accurate prediction, but the predicted clip has an overlap with the ground-truth annotation.
The woman is not shown in the last segment, and our model failed to predict this segment as positive.

%One possible reason could be that ``a scary mask'' is an ambiguous description.

%In the second example, our ASST successfully localized ``baby begins to laugh''.
%For query ``babies left hand is touching its mouth'', our ASST predicted a clip that overlaps with the ground truth annotation, but not accurate. One reason could be that our model fails to focus on the word ``left''.
%For the third query ``baby puts hand in mouth while laughing'', Our predicted clip is visually similar to the query, but the baby is not touch its mouth.
%In the third example,

% Another Example?
In the second example,
our ASST successfully detected ``we cant see the man's head in these frames'' and ``person with camera tries to hand something to the guy in white lab coat''.
%. It also detected ``the animal is eating the food'', but missed the second and the third segment.
%In the first segment, the animal is eating with no food visible in its hands.
Our ASST failed to localize ``camera tilts then corrects''. % and ``the animal approaches the cameraman aggressively''. Both descriptions are directly related to the camera (observer).
This text query is describing the camera (observer).
Events related to the observer needed to be inferred from object and camera motion, which is difficult to be directly modeled.
%The relation to the observer is difficult to model.

%TODO

\noindent{\textbf{Charades-STA}}.
We perform experiments on the Charades-STA dataset for natural language description localization in videos. Table~\ref{tab:Charades_Res} shows our results and the baseline provided by \cite{Gao_2017_ICCV}.

Our ASST achieves $37.04\%$ in Rank$@1,IoU=0.5$ and $18.04\%$ in Rank$@1,IoU=0.7$, which significantly outperforms CTRL by $15.62\%$ and $10.89\%$, respectively. In Rank$@5$s, our ASST outperforms CTRL by $9.01\%$ in $IoU=0.5$ and $11.37\%$ in $IoU=0.7$.
We also applied our model with Two-Stream visual feature as we used for the DiDeMo dataset. With Two-Stream visual feature, the performances of our model boosted to $42.72\%$ in Rank$@1,IoU=0.5$ and $24.06\%$ in Rank$@1,IoU=0.7$. This demonstrates that Two-Stream feature has a better ability to capture visual clues than C3D feature.

%SHOW SOME RESULTS
Figure~\ref{fig:result}
shows some of our localization results with C3D visual feature on the Charades-STA dataset.
In the first example, our model successfully localized ``a person sitting at a desk eating some food'' and ``person drinking from a coffee cup''. But for text query ``a person is eating at their desk'', our model predicted a long sequence with eating and drinking together, which means our model is not good enough to clearly distinguish subtle actions.
In the second example, our ASST successfully localized ``person drinks from a glass''. For text query 2 ``a smiling person runs into their garage holding a phone'', our model yields a longer prediction than the annotation.
Our model failed to detect the action ``person open the door''. This action happens at the very beginning of the video and is very short. The confidence of our predicted clip is very small, which means our model did not detect any positive clip. Better visual feature and more detailed modeling
can be used to overcome these issues.

% Some Analyses.

%---------------------------------------------------------

\begin{table*}[ht]
\footnotesize
\parbox{.45\linewidth}{

\begin{center}
\begin{tabular}{|l|c|c|c|}    
%\begin{table}[ht]
%\begin{center}
%\begin{tabular}{|l|c|c|c|}
\hline
Method & Rank$@1$ & Rank$@5$ & $mIoU$ \\
\hline\hline
%No language feature & $20.14$ & $66.10$ & $32.59$ \\
%\hline
RGB & $23.83$ & $68.25$ & $38.23$ \\
Flow & $28.37$ & $74.43$ & $44.41$ \\
Two-Stream & $30.91$ & $76.46$ & $47.30$ \\
\hline
%No RNN context & $30.81$ & $76.41$ & $45.96$ \\
%Ours & $31.05$ & $76.58$ & $47.09$ \\
RGB + Flow & $29.95$ & $77.15$ & $45.83$ \\
\hline
Fusion & $\mathbf{31.89}$ & $\mathbf{80.17}$ & $\mathbf{47.66}$ \\
\hline
\end{tabular}
\end{center}
\caption{Ablations study of input visual modalities on the DiDeMo dataset.}
\label{tab:DiDeMo_Modality}
%}
%\end{table}
%    \end{tabular}\end{center}
}
\hfill
\parbox{.45\linewidth}{
\begin{center}
\begin{tabular}{|l|c|c|c|c|}
%\begin{table}[ht]
%\begin{center}
%\begin{tabular}{|l|c|c|c|c|}
\hline
Method
& \begin{tabular}{@{}c@{}} R$@1$ \\ $IoU$=$0.5$\end{tabular}
& \begin{tabular}{@{}c@{}} R$@1$ \\ $IoU$=$0.7$\end{tabular}
& \begin{tabular}{@{}c@{}} R$@5$ \\ $IoU$=$0.5$\end{tabular}
& \begin{tabular}{@{}c@{}} R$@5$ \\ $IoU$=$0.7$\end{tabular} \\
\hline\hline
C3D & $37.04$ & $18.04$ & $68.12$ & $38.28$ \\
\hline
RGB & $31.80$ & $14.46$ & $66.69$ & $33.28$ \\
Flow & $40.48$ & $21.72$ & $70.32$ & $41.94$ \\
Two-Stream & $\mathbf{42.72}$ & $\mathbf{24.06}$ & $\mathbf{71.32}$ & $\mathbf{43.98}$ \\
\hline
\end{tabular}
\end{center}
\caption{Ablation study of input visual modalities on the Charades-STA dataset.}
\label{tab:Charades_Modality}
%\end{table}

%    \end{tabular}\end{center}
}\\
\parbox{.45\linewidth}
{
%\begin{table}[ht]
\begin{center}
\begin{tabular}{|l|c|c|c|}
\hline
Method & Rank$@1$ & Rank$@5$ & $mIoU$ \\
\hline\hline
No language & $21.10$ & $69.35$ & $33.31$ \\
\hline
Final representation layer & $26.58$ & $71.48$ & $42.40$ \\
%After context layers & $28.71$ & $75.43$ & $44.68$ \\
Last dilation layer & $29.40$ & $73.83$ & $44.68$ \\
First dilation layer & $30.12$ & $74.45$ & $46.97$ \\
%Before context layers & $29.81$ & $73.88$ & $46.02$ \\
\hline
%No RNN context & $30.81$ & $76.41$ & $45.96$ \\
%Ours & $31.05$ & $76.58$ & $47.09$ \\
Multiple feedings & $\mathbf{30.91}$ & $\mathbf{76.46}$ & $\mathbf{47.30}$ \\
\hline
\end{tabular}
\end{center}
\caption{Ablation study of modeling cross-modal local relevance on the DiDeMo dataset.}
\label{tab:DiDeMo_Merge}
%\end{table}
}
\hfill
\parbox{.45\linewidth}
{
%\begin{table}[ht]
\begin{center}
\begin{tabular}{|l|c|c|c|c|}
\hline
Method
& \begin{tabular}{@{}c@{}} R$@1$ \\ $IoU$=$0.5$\end{tabular}
& \begin{tabular}{@{}c@{}} R$@1$ \\ $IoU$=$0.7$\end{tabular}
& \begin{tabular}{@{}c@{}} R$@5$ \\ $IoU$=$0.5$\end{tabular}
& \begin{tabular}{@{}c@{}} R$@5$ \\ $IoU$=$0.7$\end{tabular} \\
\hline\hline
No language & $17.23$ & $7.50$ & $51.13$ & $22.20$ \\
\hline
Final rep. layer & $28.12$ & $13.09$ & $66.94$ & $33.84$ \\
Last dilation layer & $34.14$ & $17.34$ & $64.54$ & $35.62$ \\
First dilation layer & $35.99$ & $17.72$ & $62.39$ & $33.63$ \\
\hline
Multiple feedings & $\mathbf{37.04}$ & $\mathbf{18.04}$ & $\mathbf{68.12}$ & $\mathbf{38.28}$ \\
\hline
\end{tabular}
\end{center}
\caption{Ablation study of modeling cross-modal local relevance on the Charades-STA dataset.}
\label{tab:Charades_Merge}
%\end{table}
}
\end{table*}

\begin{table}[ht]
\footnotesize
\begin{center}
\begin{tabular}{|l|c|c|c|}
\hline
Method & Rank$@1$ & Rank$@5$ & $mIoU$ \\
\hline\hline
%MCN w/o TEF & $19.88$ & $62.39$ & $33.51$ \\
Ours & $\mathbf{30.91}$ & $76.46$ & $\mathbf{47.30}$ \\
%MCN & $27.57$ & $79.69$ & $41.70$ \\
Ours w/ TEF & $30.53$ & $\mathbf{77.34}$ & $47.14$ \\
\hline
\end{tabular}
\end{center}
\caption{Ablation study of temporal endpoint feature.}
\label{tab:DiDeMo_TEF}
\end{table}

%\subsection{Ablation studies}
\subsection{Ablation studies}
We then perform ablation studies 
%on the validation set
from the following three aspects: input visual modality, the importance of cross-modal local relevance and temporal endpoint feature. %In ablation studies, we use two-stream feature as input only.
Ablations are performed on the validation set for the DiDeMo dataset and the test set for the Charades-STA dataset.

\subsubsection{Input visual modality}
%\noindent{\textbf{Input visual modality}}.
%\textbf{Input visual modality}.
% Single model
We first perform ablation study on how the input visual modality influences our ASST's performance. We evaluate our model using RGB, optical flow and concatenation of Two-Stream features as the input visual feature.

\noindent{\textbf{DiDeMo}}.
The results of our model on the DiDeMo dataset are shown in Table~\ref{tab:DiDeMo_Modality}. 
%We trained three models which take RGB, optical flow and concatenation of two-stream features as the input visual feature. The results are shown in Table~\ref{tab:DiDeMo_Modality}.
%Within
Among these three models, RGB model achieves inferior performance. The optical flow model outperforms RGB model significantly by more than $4.54\%$ in Rank$@1$. By concatenating both RGB and optical flow input feature, Two-Stream model further improves the optical flow model by $2.54\%$ in Rank$@1$. The results show that both RGB and optical flow modality are important. Optical flow is still a good visual feature for video information modeling on tasks other than action recognition and detection. 

%Fusing
By feeding different input video data, our ASST learns relevance between natural language descriptions and video content from different aspects. Directly fusing RGB and optical flow models by a weight of $1:2$, the fused model achieves a higher performance in Rank$@5$, but lower performance in Rank$@1$ and $mIoU$ to single Two-Stream model. We then fuse all three models by a weight of $1:2:2.3$. The performance further improves around $2\%$ in all three metrics. The fusion weights are selected by
%{\color{red}{cross-validation??}}
cross-validation
%ablation study
. This fusion model is also our final model used for comparing with baselines.
%In the following ablation studies, we use single two-stream model only.

\noindent{\textbf{Charades-STA}}.
The results of our model on the Charades-STA dataset are shown in Table~\ref{tab:Charades_Modality}.
Same as the study on the DiDeMo dataset, RGB model achieves the worst performance among all models. C3D model outperforms RGB model over all metrics with better exploited spatio-temporal clues.
Optical flow captures much better motion information than RGB and C3D model, outperforms RGB model and C3D model by $8.68\%$ and $3.44\%$ on Rank$@1,IoU=0.5$, respectively. Concatenating RGB and optical flow as our input visual feature further improves Rank$@1,IoU=0.5$ to $42.72\%$.

%\noindent{\textbf{Modeling cross-modal local relevance}}.
%\textbf{Modeling cross-modal local relevance}.
\subsubsection{Modeling cross-modal local relevance}
We perform ablation study to evaluate the necessity of early fusing natural language description and video content. We use
%{\color{red}{plain???, not mentioned above.}}
plain video subnet as our baseline. This model contains no language information, and predicts clip localization directly from video content. To evaluate the effectiveness of cross-modal deep fusion and the importance of modeling cross-modal local relevance, we feed the language feature only once into video subnet via attention module. We choose three feeding positions: final representation layer, the last dilation layer and the first dilation layer. 

\noindent{\textbf{DiDeMo}}.
The results on the DiDeMo dataset are shown in Table~\ref{tab:DiDeMo_Merge}. We use Two-Stream feature as our input visual feature. In this table, the model that feeds language information on the final representation layer achieves the worst performance among all models with one-time language feature feeding. By moving language feature feeding
towards early stages, the performances increase.
Feeding language feature on the first dilation layer achieves better performances than feeding on the last dilation layer.
Both models significantly outperform the model that feeds language information on the final representation layer.

Feeding language information multiple times achieves the best performance.
The results demonstrate the importance of modeling video-language local relevance. However, feeding language feature on the final representation layer makes the model unable to build joint video-language hierarchies.
% just as we expected.

%we observe that earlier feeding achieves better performance. For earlier feeding, 

\noindent{\textbf{Charades-STA}}.
The results on the DiDeMo dataset are shown in Table~\ref{tab:DiDeMo_Merge}. We use C3D feature as our input visual feature.
We observed similar behavior of feeding position as on the DiDeMo dataset. Moving language feature feeding towards early stages improves localization accuracy on Rank$@1$s.

%On Rank$@5$s, ?????

%\noindent{\textbf{Temporal endpoint feature}}.
%\textbf{Temporal endpoint feature}.
\subsubsection{Temporal endpoint feature}
Finally, we evaluate the effect of temporal endpoint feature (TEF) proposed by Hendricks~\etal~\cite{Hendricks_2017_ICCV}. We concatenate time coordinates to the final representation layer for each frame. The results are shown in Table~\ref{tab:DiDeMo_TEF}.

TEF improves MCN significantly by $38.68\%$ relatively on Rank$@1$~\cite{Hendricks_2017_ICCV}. But TEF does not bring significant improvement to our ASST. %only improves our ASST by $1.15\%$ relatively on Rank$@5$.
One possible reason could be that during the process of visual information encoding, every element in the final representation sequence has a large receptive field. The temporal position information is encoded implicitly.

\begin{comment}

\subsection{Experiments on Charades-STA}

\subsubsection{Comparison with other models}

\subsubsection{Ablation studies}
We evaluate different model variants test set from following two aspects: input visual modality and the importance of cross-modal local relevance. %In ablation studies, we use two-stream feature as input only.

%\noindent{\textbf{Input visual modality}}.
\textbf{Input visual modality}.
We evaluate the influence on different input visual modalities. Table~\ref{tab:Charades_Modality} shows the results of different input visual features.

%\noindent{\textbf{Modeling cross-modal local relevance}}.
\textbf{Modeling cross-modal local relevance}.

\end{comment}

%\subsubsection{Training techniques}
%We first evaluate our training techniques on THUMOS 14 dataset.
%Dropout

%mOHEM

%\subsubsection{Input modal}
%RGB FLOW TWO-STREAM
%\subsubsection{tef}

\section{Conclusion and future work}
In this paper, we proposed an effective attentive sequence to sequence translator for localizing clips by natural language descriptions. We demonstrated the effectiveness of modeling vision-language information jointly. Our standalone video subnet is also an effective model for video temporal modeling. Currently, our ASST only models temporal information of videos. Rich details on video frames are ignored. For future work, we will take detailed spatial information into consideration.

{\small
\bibliographystyle{ieee}
\bibliography{egbib}

\begin{thebibliography}{10}\itemsep=-1pt

\bibitem{abadi2016tensorflow}
M.~Abadi, P.~Barham, J.~Chen, Z.~Chen, A.~Davis, J.~Dean, M.~Devin,
  S.~Ghemawat, G.~Irving, M.~Isard, et~al.
\newblock Tensorflow: A system for large-scale machine learning.
\newblock In {\em OSDI}, 2016.

\bibitem{Hendricks_2017_ICCV}
L.~Anne~Hendricks, O.~Wang, E.~Shechtman, J.~Sivic, T.~Darrell, and B.~Russell.
\newblock Localizing moments in video with natural language.
\newblock In {\em ICCV}, 2017.

\bibitem{caba2015activitynet}
F.~Caba~Heilbron, V.~Escorcia, B.~Ghanem, and J.~Carlos~Niebles.
\newblock Activitynet: A large-scale video benchmark for human activity
  understanding.
\newblock In {\em CVPR}, 2015.

\bibitem{chao2018rethinking}
Y.-W. Chao, S.~Vijayanarasimhan, B.~Seybold, D.~A. Ross, J.~Deng, and
  R.~Sukthankar.
\newblock Rethinking the faster r-cnn architecture for temporal action
  localization.
\newblock In {\em CVPR}.

\bibitem{dai2016r}
J.~Dai, Y.~Li, K.~He, and J.~Sun.
\newblock R-{fcn}: Object detection via region-based fully convolutional
  networks.
\newblock In {\em NIPS}, 2016.

\bibitem{Dai_2017_ICCV}
X.~Dai, B.~Singh, G.~Zhang, L.~S. Davis, and Y.~Qiu~Chen.
\newblock Temporal context network for activity localization in videos.
\newblock In {\em ICCV}, 2017.

\bibitem{dauphin2016language}
Y.~N. Dauphin, A.~Fan, M.~Auli, and D.~Grangier.
\newblock Language modeling with gated convolutional networks.
\newblock In {\em ICML}, 2017.

\bibitem{frome2013devise}
A.~Frome, G.~S. Corrado, J.~Shlens, S.~Bengio, J.~Dean, T.~Mikolov, et~al.
\newblock Devise: A deep visual-semantic embedding model.
\newblock In {\em NIPS}, 2013.

\bibitem{Gao_2017_ICCV}
J.~Gao, Z.~Yang, K.~Chen, C.~Sun, and R.~Nevatia.
\newblock Turn tap: Temporal unit regression network for temporal action
  proposals.
\newblock In {\em ICCV}, 2017.

\bibitem{gao2017cascaded}
J.~Gao, Z.~Yang, and R.~Nevatia.
\newblock Cascaded boundary regression for temporal action detection.
\newblock 2017.

\bibitem{gehring2017convolutional}
J.~Gehring, M.~Auli, D.~Grangier, D.~Yarats, and Y.~N. Dauphin.
\newblock Convolutional sequence to sequence learning.
\newblock In {\em ICML}, 2017.

\bibitem{girshick2015fast}
R.~Girshick.
\newblock Fast r-cnn.
\newblock In {\em ICCV}, 2015.

\bibitem{he2016deep}
K.~He, X.~Zhang, S.~Ren, and J.~Sun.
\newblock Deep residual learning for image recognition.
\newblock In {\em CVPR}, 2016.

\bibitem{hu2016natural}
R.~Hu, H.~Xu, M.~Rohrbach, J.~Feng, K.~Saenko, and T.~Darrell.
\newblock Natural language object retrieval.
\newblock In {\em CVPR}, 2016.

\bibitem{Huang_2017_CVPR}
J.~Huang, V.~Rathod, C.~Sun, M.~Zhu, A.~Korattikara, A.~Fathi, I.~Fischer,
  Z.~Wojna, Y.~Song, S.~Guadarrama, and K.~Murphy.
\newblock Speed/accuracy trade-offs for modern convolutional object detectors.
\newblock In {\em CVPR}, 2017.

\bibitem{ioffe2015batch}
S.~Ioffe and C.~Szegedy.
\newblock Batch normalization: Accelerating deep network training by reducing
  internal covariate shift.
\newblock In {\em ICML}, 2015.

\bibitem{THUMOS14}
Y.-G. Jiang, J.~Liu, A.~Roshan~Zamir, G.~Toderici, I.~Laptev, M.~Shah, and
  R.~Sukthankar.
\newblock {THUMOS} challenge: Action recognition with a large number of
  classes.
\newblock \url{http://crcv.ucf.edu/THUMOS14/}, 2014.

\bibitem{kalchbrenner2016neural}
N.~Kalchbrenner, L.~Espeholt, K.~Simonyan, A.~v.~d. Oord, A.~Graves, and
  K.~Kavukcuoglu.
\newblock Neural machine translation in linear time.
\newblock {\em arXiv preprint arXiv:1610.10099}, 2016.

\bibitem{karpathy2014large}
A.~Karpathy, G.~Toderici, S.~Shetty, T.~Leung, R.~Sukthankar, and L.~Fei-Fei.
\newblock Large-scale video classification with convolutional neural networks.
\newblock In {\em CVPR}.

\bibitem{kingma2014adam}
D.~Kingma and J.~Ba.
\newblock Adam: A method for stochastic optimization.
\newblock In {\em ICLR}, 2015.

\bibitem{Lea_2017_CVPR}
C.~Lea, M.~D. Flynn, R.~Vidal, A.~Reiter, and G.~D. Hager.
\newblock Temporal convolutional networks for action segmentation and
  detection.
\newblock In {\em CVPR}, 2017.

\bibitem{Lin:2017:SST:3123266.3123343}
T.~Lin, X.~Zhao, and Z.~Shou.
\newblock Single shot temporal action detection.
\newblock In {\em ACM MM}, 2017.

\bibitem{lin2016feature}
T.-Y. Lin, P.~Doll{\'a}r, R.~Girshick, K.~He, B.~Hariharan, and S.~Belongie.
\newblock Feature pyramid networks for object detection.
\newblock In {\em CVPR}, 2017.

\bibitem{Lin_2017_ICCV}
T.-Y. Lin, P.~Goyal, R.~Girshick, K.~He, and P.~Dollar.
\newblock Focal loss for dense object detection.
\newblock In {\em ICCV}, 2017.

\bibitem{liu2017referring}
J.~Liu, L.~Wang, M.-H. Yang, et~al.
\newblock Referring expression generation and comprehension via attributes.
\newblock In {\em CVPR}.

\bibitem{liu2016ssd}
W.~Liu, D.~Anguelov, D.~Erhan, C.~Szegedy, S.~Reed, C.-Y. Fu, and A.~C. Berg.
\newblock Ssd: Single shot multibox detector.
\newblock In {\em ECCV}, 2016.

\bibitem{mao2016generation}
J.~Mao, J.~Huang, A.~Toshev, O.~Camburu, A.~L. Yuille, and K.~Murphy.
\newblock Generation and comprehension of unambiguous object descriptions.
\newblock In {\em CVPR}, 2016.

\bibitem{oord2016wavenet}
A.~v.~d. Oord, S.~Dieleman, H.~Zen, K.~Simonyan, O.~Vinyals, A.~Graves,
  N.~Kalchbrenner, A.~Senior, and K.~Kavukcuoglu.
\newblock Wavenet: A generative model for raw audio.
\newblock {\em arXiv preprint arXiv:1609.03499}, 2016.

\bibitem{pantofaru2017ava}
C.~Pantofaru, C.~Sun, C.~Gu, C.~Schmid, D.~Ross, G.~Toderici, J.~Malik,
  R.~Sukthankar, S.~Vijayanarasimhan, S.~Ricco, et~al.
\newblock Ava: A video dataset of spatio-temporally localized atomic visual
  actions.
\newblock 2017.

\bibitem{pennington2014glove}
J.~Pennington, R.~Socher, and C.~Manning.
\newblock Glove: Global vectors for word representation.
\newblock In {\em EMNLP}, 2014.

\bibitem{ren2015faster}
S.~Ren, K.~He, R.~Girshick, and J.~Sun.
\newblock Faster {R-CNN}: Towards real-time object detection with region
  proposal networks.
\newblock In {\em NIPS}, 2015.

\bibitem{Richard_2016_CVPR}
A.~Richard and J.~Gall.
\newblock Temporal action detection using a statistical language model.
\newblock In {\em CVPR}, 2016.

\bibitem{rohrbach2016grounding}
A.~Rohrbach, M.~Rohrbach, R.~Hu, T.~Darrell, and B.~Schiele.
\newblock Grounding of textual phrases in images by reconstruction.
\newblock In {\em ECCV}, 2016.

\bibitem{russakovsky2015imagenet}
O.~Russakovsky, J.~Deng, H.~Su, J.~Krause, S.~Satheesh, S.~Ma, Z.~Huang,
  A.~Karpathy, A.~Khosla, M.~Bernstein, et~al.
\newblock Imagenet large scale visual recognition challenge.
\newblock {\em IJCV}, 115(3):211--252, 2015.

\bibitem{cdc_shou_cvpr17}
Z.~Shou, J.~Chan, A.~Zareian, K.~Miyazawa, and S.-F. Chang.
\newblock Cdc: Convolutional-de-convolutional networks for precise temporal
  action localization in untrimmed videos.
\newblock In {\em CVPR}, 2017.

\bibitem{Shou_2016_CVPR}
Z.~Shou, D.~Wang, and S.-F. Chang.
\newblock Temporal action localization in untrimmed videos via multi-stage
  cnns.
\newblock In {\em CVPR}, 2016.

\bibitem{shrivastava2016beyond}
A.~Shrivastava, R.~Sukthankar, J.~Malik, and A.~Gupta.
\newblock Beyond skip connections: Top-down modulation for object detection.
\newblock {\em arXiv preprint arXiv:1612.06851}, 2016.

\bibitem{sigurdsson2016hollywood}
G.~A. Sigurdsson, G.~Varol, X.~Wang, A.~Farhadi, I.~Laptev, and A.~Gupta.
\newblock Hollywood in homes: Crowdsourcing data collection for activity
  understanding.
\newblock In {\em ECCV}, 2016.

\bibitem{simonyan2014two}
K.~Simonyan and A.~Zisserman.
\newblock Two-stream convolutional networks for action recognition in videos.
\newblock In {\em NIPS}, 2014.

\bibitem{simonyan2014very}
K.~Simonyan and A.~Zisserman.
\newblock Very deep convolutional networks for large-scale image recognition.
\newblock In {\em ICLR}, 2015.

\bibitem{UCF101}
K.~Soomro, A.~Roshan~Zamir, and M.~Shah.
\newblock {UCF101}: A dataset of 101 human actions classes from videos in the
  wild.
\newblock In {\em CRCV-TR-12-01}, 2012.

\bibitem{tran2015learning}
D.~Tran, L.~Bourdev, R.~Fergus, L.~Torresani, and M.~Paluri.
\newblock Learning spatiotemporal features with 3d convolutional networks.
\newblock In {\em ICCV}, 2015.

\bibitem{venugopalan15iccv}
S.~Venugopalan, M.~Rohrbach, J.~Donahue, R.~Mooney, T.~Darrell, and K.~Saenko.
\newblock Sequence to sequence -- video to text.
\newblock In {\em ICCV}, 2015.

\bibitem{wang2015towards}
L.~Wang, Y.~Xiong, Z.~Wang, and Y.~Qiao.
\newblock Towards good practices for very deep two-stream convnets.
\newblock {\em arXiv preprint arXiv:1507.02159}, 2015.

\bibitem{wang2016temporal}
L.~Wang, Y.~Xiong, Z.~Wang, Y.~Qiao, D.~Lin, X.~Tang, and L.~Van~Gool.
\newblock Temporal segment networks: Towards good practices for deep action
  recognition.
\newblock In {\em ECCV}, 2016.

\bibitem{wu2016google}
Y.~Wu, M.~Schuster, Z.~Chen, Q.~V. Le, M.~Norouzi, W.~Macherey, M.~Krikun,
  Y.~Cao, Q.~Gao, K.~Macherey, et~al.
\newblock Google's neural machine translation system: Bridging the gap between
  human and machine translation.
\newblock {\em arXiv preprint arXiv:1609.08144}, 2016.

\bibitem{Xu_2017_ICCV}
H.~Xu, A.~Das, and K.~Saenko.
\newblock R-c3d: Region convolutional 3d network for temporal activity
  detection.
\newblock In {\em ICCV}, 2017.

\bibitem{yu2015multi}
F.~Yu and V.~Koltun.
\newblock Multi-scale context aggregation by dilated convolutions.
\newblock In {\em ICLR}, 2016.

\bibitem{Zhao_2017_ICCV}
Y.~Zhao, Y.~Xiong, L.~Wang, Z.~Wu, X.~Tang, and D.~Lin.
\newblock Temporal action detection with structured segment networks.
\newblock In {\em ICCV}, 2017.

\end{thebibliography}
}

\clearpage

\section*{Appendix}

%\section*{Appendix}
To validate our video subnet's ability for temporal visual information modeling,
we perform experiments for a relevant task, ``action detection'', on the THUMOS 14 dataset~\cite{THUMOS14}.

THUMOS 14 is a video dataset for action recognition and detection. There are $101$ classes for action recognition task and $20$ classes for detection task. For action detection task, usually untrimmed videos from validation set and testing set are used as training data and testing data. There are $1,010$ videos in the validation set and $1,574$ in the testing set respectively. Only $200$ and $213$ videos in these two sets contain annotations within the $20$ action classes. We only use these $413$ videos for training and evaluation following most recent literatures. The metrics for THUMOS 14 action detection task are $mAP$s (mean average precision) on different $IoU$s (Intersection over Union).

%This task contains no natural language descriptions. Interested action clips are annotated by $20$ pre-defined class labels.
We use the detection model of standalone video subnet as our model. 
This model contains four dilation layers, six squeezing layers and six expansion layers, which is the same as we used for localizing clips by natural language descriptions task.
Language subnet and attention module are removed for action detection task. Clip samplers with $21$-way classifiers ($20$ classes and $1$ background class) and temporal coordinate regression are built upon the final representation sequence.
For a fair comparison, we use Inception-BN trained on ImageNet training set as our RGB network, and Inception-BN trained on UCF101~\cite{UCF101} split $1$ training set~\cite{wang2016temporal} as our optical flow network, following \cite{Zhao_2017_ICCV}.

We use Stochastic Gradient Descent (SGD) with batch size $128$ to train our model. The learning rate starts with $1e-2$, and multiplies $0.9$ every $1,000$ iterations. During training, we sample clips with positive:negative ratio $1:5$. For inference, we slide our model along temporal domain for every testing video, followed by Non-Maximum-Suppression (NMS) with threshold $0.3$ as post processing.

We show our action detection performances on the THUMOS 14 dataset in Table~\ref{tab:THUMOS14}.
%Some of recent literatures also showed their results at $IoU$ $0.6$ and $0.7$. We also show our results at $IoU$ $0.6$ and $0.7$ for a full comparison.
%We report our results at $IoU$ $0.3$ to $0.7$ due to limit of space.
We report our results at $IoU$ $0.3$ to $0.7$ following Shou~\etal\cite{cdc_shou_cvpr17}.

\begin{table}[ht]
\footnotesize
\begin{center}
\begin{tabular}{|l|c|c|c|c|c|}
%\hline
%\multicolumn{8}{c}{Performances on THUMOS14, mAP@IoU $\alpha$} \\
\hline
Method & 0.3 & 0.4 & 0.5 & 0.6 & 0.7 \\
\hline\hline
Richard~\etal~\cite{Richard_2016_CVPR} & $30.0$ & $23.2$ & $15.2$ & $-$ & $-$ \\
Shou~\etal~\cite{Shou_2016_CVPR} & $36.3$ & $28.7$ & $19.0$ & $10.3$ & $5.3$ \\
Shou~\etal~\cite{cdc_shou_cvpr17} & $40.1$ & $29.4$ & $23.3$ & $13.1$ & $7.9$ \\
Lin~\etal~\cite{Lin:2017:SST:3123266.3123343} & $43.0$ & $35.0$ & $24.6$ & $-$ & $-$ \\
Dai~\etal~\cite{Dai_2017_ICCV} & $-$ & $33.3$ & $25.6$ & $15.9$ & $9.0$ \\
Gao~\etal~\cite{Gao_2017_ICCV} & $44.1$ & $34.9$ & $25.6$ & $-$ & $-$ \\
Xu~\etal~\cite{Xu_2017_ICCV} & $44.8$ & $35.6$ & $28.9$ & $-$ & $-$ \\
Zhao~\etal~\cite{Zhao_2017_ICCV} & $51.9$ & $41.0$ & $29.8$ & $19.6$ & $10.7$ \\
Gao~\etal~\cite{gao2017cascaded} & $50.1$ & $41.3$ & $31.0$ & $19.1$ & $9.9$ \\
%Bai~\etal~\cite{bai2018contextual} & $54.7$ & $48.2$ & $40.0$ & - & - \\
Chao~\etal~\cite{chao2018rethinking} & $53.2$ & $48.5$ & $42.8$ & $33.8$ & $20.8$ \\

\hline
%Ours & 56.29 & 55.55 & \textbf{53.94} & \textbf{50.41} & \textbf{44.95} & \textbf{37.49} & \textbf{27.35} \\
%Ours & $56.3$ & $55.6$ & $\mathbf{53.9}$ & $\mathbf{50.4}$ & $\mathbf{45.0}$ & $\mathbf{37.5}$ & $\mathbf{27.4}$ \\
%0.5665	0.5575	0.5335	0.51	0.4555	0.3687	0.2612
%Ours ($200$ training videos) & $56.7$ & $55.8$ & $\mathbf{53.4}$ & $\mathbf{51.0}$ & $\mathbf{45.6}$ & $\mathbf{36.9}$ & $\mathbf{26.1}$ \\ % 5e-3 mOHEM 20000 old test
%Ours & $56.7$ & $55.8$ & $\mathbf{53.4}$ & $\mathbf{51.0}$ & $\mathbf{45.6}$ & $\mathbf{36.9}$ & $\mathbf{26.1}$ \\
%0.57044	0.565223	0.539125	0.502734	0.439835	0.353739	0.236448
%Ours & $57.0$ & $56.5$ & $\mathbf{53.9}$ & $\mathbf{50.3}$ & $\mathbf{44.0}$ & $\mathbf{35.4}$ & $\mathbf{23.6}$ \\ % 5e-3 mOHEM 20000
%0.577673	0.571917	0.553698	0.513192	0.448076	0.350251	0.225191
%Ours & $57.8$ & $57.2$ & $\mathbf{55.4}$ & $\mathbf{51.3}$ & $\mathbf{44.8}$ & $\mathbf{35.0}$ & $\mathbf{22.5}$ \\ % 5e-3 rOHEM 15000
%0.580376	0.569213	0.543361	0.502338	0.447252	0.36089	0.230125
%Ours & $58.0$ & $56.9$ & $\mathbf{54.3}$ & $\mathbf{50.2}$ & $\mathbf{44.7}$ & $\mathbf{36.1}$ & $\mathbf{23.0}$ \\ % 5e-3 mOHEM 13000
%Ours & $\mathbf{54.3}$ & $\mathbf{50.2}$ & $\mathbf{44.7}$ & $\mathbf{36.1}$ & $\mathbf{23.0}$ \\ % 5e-3 mOHEM 13000
%0.576878	0.56941	0.547648	0.505152	0.450518	0.360867	0.233621
Ours & $\mathbf{54.9}$ & $\mathbf{50.3}$ & $\mathbf{43.7}$ & $\mathbf{34.4}$ & $\mathbf{22.2}$ \\ % 1e-2 neg5 20000
%0.581431	0.569852	0.54855	0.503035	0.437257	0.343657	0.222381
%0.566099	0.554099	0.537278	0.503393	0.436262	0.357726	0.243617

%0.593169	0.585946	0.554702	0.513554	0.453901	0.364523	0.25451

\hline
\end{tabular}
\end{center}
\caption{Action detection results on the THUMOS 14 dataset (in percentage). The $IoU$ threshold used in evaluation varies from $0.3$ to $0.7$. - indicates the results in corresponding papers are unavailable.}
\label{tab:THUMOS14}
\end{table}

%The most frequently used metric in recent papers for action detection is $mAP@IoU~0.5$. 
Our model is able to outperform state-of-the-art approaches over all metrics. 
Our method outperforms earlier approaches~\cite{Zhao_2017_ICCV, gao2017cascaded} by a large margin of more than $10\%$ on $mAP@IoU~0.5$. Comparing to very recent approaches~\cite{chao2018rethinking}, our model still achieves very competitive results of around $1\%$ improvement over all metrics.

%On $mAP@IoU~0.5$, our model outperforms previous approaches by a large margin. 
%On other metrics, our model also achieves highly competitive results. 
%Our model outperforms state-of-the-art methods by a notable margin of $15.3\%$.
%Zhao~\etal~\cite{Zhao_2017_ICCV} achieved a good performance among all recent methods by building a hard coded context region.
The video subnet encodes temporal visual context via multiple convolutional layers, which learns deep contextual temporal structure with multiple granularities, and is more capable of capturing video content at different scales, thereby achieving a better performance.
The experimental results for action detection on the THUMOS 14 dataset show that our video subnet is able to capture good temporal visual structures for clip localization.
%the ability of our video subnet to model temporal visual information. 
The ability of modeling temporal visual structures is an essential ingredient of our ASST, enables our ASST to better understand visual information and linguistic information jointly.

%the video context information are already encoded into the model implicitly. 
%Zhao~\etal achieved a superior performance at $IoU$ $0.1$ and $0.2$. But others $mAP$ decreases fast as $IoU$ increases, while ours decreases much slower.

%Comparing to other methods, 

%This is because our model makes more accurate predictions. We show our prediction in figure XXX.
% SHOW EXAMPLE?

% Our model can outperform most of recent results. Especially on larger $IoU$s, we can outperform state-of-the-art results by a large margin. 
% As IoU increases, our performance decreases slower than other methods. Which means, for predicted clips, our model is able to predict more accurate location estimation.

%Zhao~\etal~\cite{Zhao_2017_ICCV} used a sophisticated merging method, so we are unable to outperform them.

% The experimental results show that our visual subnet is good enough to encode video information.

\end{document}